\documentclass{llncs}
\usepackage{makeidx}  
\usepackage{epsbox}
\begin{document}

\mainmatter              
\title{Meaning Sort\\
--- Three examples: dictionary construction, tagged corpus 
     construction, and  information presentation system ---}
\titlerunning{Meaning Sort}  
%
\author{Masaki Murata \and Kyoko Kanzaki \and Kiyotaka Uchimoto \and \\Qing Ma \and Hitoshi Isahara}

\authorrunning{Masaki Murata et al.}   
%
\tocauthor{Masaki Murata(Communications Research Laboratory, MPT),
Kyoko Kanzaki(Communications Research Laboratory, MPT),
Kiyotaka Uchimoto(Communications Research Laboratory, MPT),
Qing Ma(Communications Research Laboratory, MPT),
Hitoshi Isahara(Communications Research Laboratory, MPT)}

\institute{Communications Research Laboratory, MPT,\\ 2-2-2 Hikaridai, Seika-cho, Soraku-gun, Kyoto, 619-0289, Japan,\\
\email{\{murata,kanzaki,uchimoto,qma,isahara\}@crl.go.jp},\\ WWW home page:
\texttt{http://www-karc.crl.go.jp/ips/murata}}

\maketitle              

\def\Noun#1{{\sf NP #1}}
\def\Verb#1{{\sf Verb #1}}

\def\gyouji{an event}
\def\ningen{Human}
\def\koushitsu{the Imperial Household}
\def\oushitsu{a Royal family}
\def\kanmin{a government official}
\def\iemoto{the head of a school}
\def\soshiki{Organization}
\def\zenkoku{the whole country}
\def\nouson{an agricultural village}
\def\ken{a prefecture}
\def\nihon{Japan}
\def\soren{the Soviet Union}
\def\tera{a temple}
\def\gakkou{a school}
\def\gakuen{a campus}
\def\bokou{an alma mater}
\def\katsudou{Action}
\def\iwai{a celebration}
\def\kourei{an established custom}
\def\koushiki{a formal style}
\def\shuunin{to take up one's post}
\def\matsuri{a festival}

\def\suuryou{Quantity}
\def\kankei{Relation}
\def\doubutsu{Animal}
\def\seisanbutsu{Product}
\def\doubutsunobubun{Part of a living thing}
\def\taibu{Part of a living thing}
\def\shokubutsu{Plant}
\def\shizenbutsu{Nature}
\def\kuukanhougaku{Location}
\def\kuukan{Location}
\def\jikan{Time}
\def\genshomeishi{Phenomenon}
\def\genshou{Phenomenon}
\def\tyuushoukankei{Abstract relation}
\def\ningenkatsudou{Human activity}
\def\ningenkatsudou{Human activity}
\def\sonohoka{Other}

\def\shurui{Style}
\def\taishoubutsu{Object}
\def\keijou{Depth}
\def\saizu{Size}
\def\zaishitsu{Material}

\def\hirautsuwa{{\it utsuwa} (a container)}
\def\utsuwa{{\it utsuwa} (a container)}
\def\ishiwan{{\it wan1} (a ceramic bowl)}
\def\kiwan{{\it wan2} (a wooden bowl)}
\def\yunomi{{\it yunomi} (a Japanese teacup)}
\def\sara{{\it sara} (a plate)}

\def\wakeishiki{Oriental}
\def\ryokutya{Japanese tea}
\def\hukai{deep}
\def\asai{shallow}
\def\toujikeishiki{ceramic}
\def\mokuseino{wooden}

\def\migiueni{continued in the right-hand column}

\begin{abstract}
It is often useful to sort words into an order 
that reflects relations among their meanings 
as obtained by using a thesaurus. 
In this paper, we introduce a method of arranging words semantically 
by using several types of `{\sf is-a}' thesauri 
and a multi-dimensional thesaurus. 
We also describe three major applications where 
a meaning sort is useful and 
show the effectiveness of a meaning sort. 
Since there is no doubt that 
a word list in meaning-order is easier to use 
than a word list in some random order, 
a meaning sort, which can easily produce 
a word list in meaning-order, 
must be useful and effective. 
\end{abstract}
%

\section{Using Msort}
\label{sec:msort}

Arranging words in an order that is based on 
their meanings is called 
a meaning sort (Msort). 
The Msort is a method of arranging words by their meanings 
rather than alphabetically. 
The method used 
to list the meanings 
is described in the next section.

For example, 
suppose we obtain the following data in a research project:\footnote{
We actually obtained this data from the EDR co-occurrence dictionary \cite{EDR93e}. }\\

\begin{center}
{
  \begin{tabular}[h]{p{10cm}}
\fbox{\gyouji} \\
\tera, \koushiki, \bokou, \shuunin, \koushitsu, \gakuen, \nihon, \soren, \zenkoku, \nouson, 
\ken, \gakkou, \matsuri, \iemoto, \kourei, \kanmin, \iwai, \oushitsu\\
  \end{tabular}
}
\end{center}

This is a list of noun phrases ({\sf NP}s), 
each followed by the word {\it gyoji} (\gyouji) 
in the form \Noun{X} {\it no gyoji} (\gyouji \ of \Noun{X}) 
in Japanese. 
To find the most useful way to examine the list, 
we first arrange the NPs alphabetically:\\

\begin{center}
{
  \begin{tabular}[h]{p{10cm}}
\nouson, \bokou, \gakuen, \iwai, \kourei, \matsuri, \koushiki, \kanmin, \iemoto, 
\koushitsu, \nihon, \ken, \oushitsu, \gakkou, \soren, \shuunin, \tera, \zenkoku 
\end{tabular}
}
\end{center}

This list is not easy to use, 
so we next arrange the NPs by frequency of appearance: \\

\begin{center}
{
  \begin{tabular}[h]{p{10cm}}
\kourei, \gakkou, \koushiki, \nihon, \ken, \zenkoku, \tera, \nouson, \oushitsu, \soren, 
\matsuri, \gakuen, \shuunin, \iwai, \bokou, \koushitsu, \kanmin, \iemoto
\end{tabular}
}
\end{center}

Yet, even arranged this way, it is too difficult 
to use the list. 

We then use an Msort to arrange the NPs semantically, 
by using following categories: \ningen, \soshiki, and \katsudou: \\

\begin{center}
{
  \begin{tabular}[h]{l@{ }p{7cm}}
(\ningen) & \koushitsu, \oushitsu, \kanmin, \iemoto \\
(\soshiki) &\zenkoku, \nouson, \ken, \nihon, \soren, \tera, \gakkou, \gakuen, \bokou \\
(\katsudou) &\iwai, \kourei, \koushiki, \shuunin, \matsuri\\
\end{tabular}
}
\end{center}

This list is much easier to use than 
a listing in alphabetical or frequency order. 
Note that the words in each line are also arranged in an order 
that reflects relations among their meanings. 
For example, 
{\it \nihon} \ and {\it \soren} are listed side by side, 
as are {\it \gakkou}, {\it \gakuen}, \ and {\it \bokou}. 

Although the list shows a variety of events, 
we can see at a glance that 
some are events related to certain special persons, 
and some are events related to a certain organization, 
and the others are miscellaneous forms of events. 

The Msort is also applicable to other situations 
as described in later sections. 
The Msort enables users to more easily and efficiently 
recognize and examine various types of problems.

\begin{table}[t]
    \caption{Modified BGH category numbers}
    \label{tab:bunrui_code_change}
  \begin{center}
\begin{tabular}[c]{|l|l|l|}\hline
     Semantic marker   &  Original        & Modified\\
                       &  code           &  code\\\hline
     Animal            &  [1-3]56                & 511\\
     Human             &  12[0-4]            & 52[0-4]\\
     Organization      &  [1-3]2[5-8]        & 53[5-8]\\
     Products          &  [1-3]4[0-9]            & 61[0-9]\\
     Part of a living thing &  [1-3]57                & 621\\
     Plant             &  [1-3]55                & 631\\
     Nature            &  [1-3]52                & 641\\
     Location          &  [1-3]17                & 657\\
     Quantity          &  [1-3]19                & 711\\
     Time              &  [1-3]16                & 811\\
     Phenomenon        &  [1-3]5[01]            & 91[12]\\
     Abstract relation &  [1-3]1[0-58]           & aa[0-58]\\
     Human activity    &  [1-3]58,[1-3]3[0-8]        & ab[0-9]\\
     Other             &  4                  & d\\\hline
\end{tabular}
\end{center}
\end{table}

\begin{table*}[t]
  \caption{An example of the Msort process}
    \label{tab:bgh_rei3}

\addtocounter{table}{-1}

    \begin{center}

(a) \ Examples with BGH category numbers

\begin{tabular}[t]{|@{ }l@{\hspace{0.5cm}}l@{ }|}\hline
5363005022 & \tera    \\
5363005021 & \tera    \\
ab18207012 & \koushiki  \\
ab21509016 & \koushiki  \\
aa11011014 & \koushiki  \\
ab70004013 & \koushiki  \\
5363013015 & \bokou  \\
ab41201016 & \shuunin  \\
5210007021 & \koushitsu  \\
5363010015 & \gakuen  \\
5359001012 & \nihon  \\
5359004192 & \soren  \\
\multicolumn{2}{|l|}{\migiueni}\\\hline
\end{tabular}
\begin{tabular}[t]{|@{ }l@{\hspace{0.5cm}}l@{ }|}\hline
7118007013 & \zenkoku\\
5353007012 & \zenkoku\\
5354006033 & \nouson\\
5355004017 & \ken\\
5363010012 & \gakkou\\
ab46002012 & \matsuri\\
5241023012 & \iemoto\\
ab18205021 & \kourei\\
5233004015 & \kanmin\\
5241101061 & \kanmin\\
ab14308013 & \iwai\\
ab46019012 & \iwai\\
5210007022 & \oushitsu\\\hline
\end{tabular}

\end{center}

\end{table*}

\begin{table*}[p]
  \caption{Example of the Msort process}
    \label{tab:bgh_rei3}

    \begin{center}

\begin{tabular}[c]{c@{ }c@{ }c}
\begin{tabular}[t]{|@{ }l@{\hspace{0.5cm}}l@{}|}
\multicolumn{2}{@{}l}{
  \begin{tabular}[t]{@{}l@{ }p{4cm}}
(b) & Adding semantic markers for divisions\\
\end{tabular}}\\\hline
5100000000 & (\doubutsu)\\
5200000000 & (\ningen)\\
5300000000 & (\soshiki)\\
6100000000 & (\seisanbutsu)\\
6200000000 & (\doubutsunobubun)\\
6300000000 & (\shokubutsu)\\
6400000000 & (\shizenbutsu)\\
6500000000 & (\kuukanhougaku)\\
7100000000 & (\suuryou)\\
8100000000 & (\jikan)\\
9100000000 & (\genshomeishi)\\
aa00000000 & (\tyuushoukankei)\\
ab00000000 & (\ningenkatsudou)\\
d000000000 & (\sonohoka)\\
5363005022 & \tera\\
5363005021 & \tera\\
ab18207012 & \koushiki\\
ab21509016 & \koushiki\\
aa11011014 & \koushiki\\
ab70004013 & \koushiki\\
5363013015 & \bokou\\
ab41201016 & \shuunin\\
5210007021 & \koushitsu\\
5363010015 & \gakuen\\
5359001012 & \nihon\\
5359004192 & \soren\\
7118007013 & \zenkoku\\
5353007012 & \zenkoku\\
5354006033 & \nouson\\
5355004017 & \ken\\
5363010012 & \gakkou\\
ab46002012 & \matsuri\\
5241023012 & \iemoto\\
ab18205021 & \kourei\\
5233004015 & \kanmin\\
5241101061 & \kanmin\\
ab14308013 & \iwai\\
ab46019012 & \iwai\\
5210007022 & \oushitsu\\\hline
\end{tabular}
&
\begin{tabular}[t]{|@{ }l@{\hspace{0.5cm}}l@{}|}
\multicolumn{2}{@{}l}{
  \begin{tabular}[t]{@{}l@{ }p{4.5cm}}
(c) & Arranging elements in the order of their category number\\
\end{tabular}}\\\hline
5100000000 & (\doubutsu)\\
5200000000 & (\ningen)\\
5210007021 & \koushitsu\\
5210007022 & \oushitsu\\
5233004015 & \kanmin\\
5241023012 & \iemoto\\
5241101061 & \kanmin\\
5300000000 & (\soshiki)\\
5353007012 & \zenkoku\\
5354006033 & \nouson\\
5355004017 & \ken\\
5359001012 & \nihon\\
5359004192 & \soren\\
5363005021 & \tera\\
5363005022 & \tera\\
5363010012 & \gakkou\\
5363010015 & \gakuen\\
5363013015 & \bokou\\
6100000000 & (\seisanbutsu)\\
6200000000 & (\doubutsunobubun)\\
6300000000 & (\shokubutsu)\\
6400000000 & (\shizenbutsu)\\
6500000000 & (\kuukanhougaku)\\
7100000000 & (\suuryou)\\
7118007013 & \zenkoku\\
8100000000 & (\jikan)\\
9100000000 & (\genshomeishi)\\
aa00000000 & (\tyuushoukankei)\\
aa11011014 & \koushiki\\
ab00000000 & (\ningenkatsudou)\\
ab14308013 & \iwai\\
ab18205021 & \kourei\\
ab18207012 & \koushiki\\
ab21509016 & \koushiki\\
ab41201016 & \shuunin\\
ab46002012 & \matsuri\\
ab46019012 & \iwai\\
ab70004013 & \koushiki\\
d000000000 & (\sonohoka)\\\hline
\end{tabular}
\end{tabular}
      
    \end{center}

\end{table*}

\section{Implementing Msort}
\label{sec:implement}

To sort words in an order that reflects relations among their meanings, 
we first need to determine an order for the meanings. 
The Japanese thesaurus {\it Bunrui Goi Hyou} \cite{NLRI64ae}, 
an `{\sf is-a}' hierarchical thesaurus, is useful for this. 
We refer to it as {\it BGH}. 
In BGH, each word has {\it a category number}. 
In the electronic version of BGH, 
each word has a 10-digit category number that 
indicates 
seven levels of the `{\sf is-a}' hierarchy. 
The top five levels are expressed by 
the first five digits, 
the sixth level is expressed 
by the next two digits, and 
the last level is expressed by the last three digits. 
(Although we have used BGH, Msort can also be used with 
other thesauri in other languages.)

The easiest way of implementing Msort is to 
arrange words in order of their category numbers. 
However, only arranging words semantically does not produce a convenient result. 
If the items arranged are numbers, the order is clear, 
but there is no clear order for words. 
It is thus convenient to insert a mark, 
as a kind of bookmark, in certain places. 
We used semantic markers 
such as {\it Human}, {\it Organization} and {\it Action} as bookmarks.

These markers were created by 
combining nominal semantic markers 
in the IPAL verbal dictionary \cite{IPALe} 
with the BGH classification system. 
Table \ref{tab:bunrui_code_change} shows 
the modified category numbers 
obtained by integrating these new markers 
with the BGH codes. 
The first three digits of each category number have been changed. 
For example, the notation 
[1-3]56 and 511 in the first line 
means that 
when the first three digits of the category number 
are 156, 256, or 356, 
those digits will be changed to 511. 
([1-3] means 1, 2, or 3.) 

The process of using an Msort is explained 
by applying it to the data set listed in 
Section \ref{sec:msort}, 
obtained by the word {\it gyoji} (\gyouji), 
as follows: 

\begin{table}[t]
  \caption{Definitions of concepts 
from the top node to the node of the term ``\bokou''}
  \label{tab:def_concepts_bokou}
\begin{center}
{\begin{tabular}[c]{|l|}\hline
concept\\
agent\\
autonomous being\\
organization\\
educational organization\\
an organization to provide education, called a school\\
a school at which a person was or is a student\\\hline
\end{tabular}\\
}
\end{center}
\end{table}

\begin{table}[t]
  \caption{Results of an Msort using the BGH thesaurus}
  \label{tab:last_rei}
  \begin{center}
\begin{tabular}[c]{|l@{ }p{9.5cm}|}\hline
(\ningen) & \koushitsu, \oushitsu, \kanmin, \iemoto \\
(\soshiki) & \zenkoku, \nouson, \ken, \nihon, \soren, \tera, \gakkou, \gakuen, \bokou \\
(\suuryou) & \zenkoku \\
(\kankei) & \koushiki \\
(\katsudou) & \iwai, \kourei, \koushiki, \shuunin, \matsuri \\\hline
\end{tabular}
\end{center}
\end{table}

\begin{table*}[t]
  \caption{Results of an Msort using the EDR dictionary}
  \label{tab:EDR_last_rei}
  \begin{center}
\begin{tabular}[c]{|l@{ }c@{ }l@{ }c@{ }lp{6cm}|}\hline
(concept & : & agent & : & autonomous being) & \gakkou, \gakuen, \bokou, \tera, \ken, \soren, \nihon, \oushitsu, \koushitsu, \iemoto, \kanmin\\
(concept & : & agent & : & human being) & \tera, \ken, \iemoto, \kanmin\\
(concept & : & event & : & action) & \iwai, \shuunin\\
(concept & : & event & : & phenomenon) & \matsuri\\
(concept & : & matter & : & event) & \matsuri, \kourei, \iwai\\
(concept & : & matter & : & thing) & \tera, \gakkou, \ken, \iemoto, \kanmin, \iwai, \koushiki\\
(concept & : & space & : & location) & \tera, \gakkou, \zenkoku, \ken, \nouson, \soren, \nihon\\\hline
\end{tabular}
\end{center}
\end{table*}

\begin{enumerate}
\item 
  Firstly, we give each word a new category number 
  according to the transformation shown in Table \ref{tab:bunrui_code_change}, 
  to obtain the results 
  shown in Table \ref{tab:bgh_rei3}(a). 
  {\it A temple} occurs twice, and 
  {\it \koushiki} \ occurs four times. 
  This indicates that both {\it \tera} \ and {\it \koushiki} \ 
  have multiple meanings. 
  In the BGH thesaurus, {\it \tera} \ is defined as having two meanings, and 
  {\it \koushiki} \ is defined as having four meanings. 

\item 
  We then add semantic markers to the set of words 
  in Table \ref{tab:bgh_rei3}(a) 
  to get the results shown in Table \ref{tab:bgh_rei3}(b). 
\item 
  Next, we arrange the items in Table \ref{tab:bgh_rei3}(b) 
  in the order of their category numbers 
  to get the results shown in Table \ref{tab:bgh_rei3}(c). 
\item 
  Finally, we convert the data into a form that is easier to use. 
  For example, 
  when we delete the category numbers, 
  redundant words with the same semantic marker in a line, 
  and semantic markers to which no words correspond, 
  we obtain the data shown in Table \ref{tab:last_rei}. 
\end{enumerate}

This data is much easier to use 
than the data shown in the other tables. 

\section{Msort using different dictionaries}

\subsection{Msort using a different `{\sf is-a}' thesaurus}

In Section \ref{sec:implement} 
we described the implementation of an Msort using the BGH thesaurus. 
This is the most suitable `is-a' thesaurus for an Msort 
because each word which contains is assigned a category number. 
This section examines whether an Msort can be used 
with an `{\sf is-a}' hierarchical thesaurus which has no category numbers, 
such as the EDR dictionary \cite{EDR93e}. 

It is useful to consider 
the definition sentence of the concept in each node 
of an {\sf is-a} thesaurus as the number of the level. 
If we do this, it is not necessary to create a new number. 
For example, 
the definitions of concepts 
from the top node to the node of the term ``\bokou'' 
are as shown in Table \ref{tab:def_concepts_bokou}. 

When we do a meaning sort using 
the EDR dictionary, 
we only have to consider 
the connections of the hierarchy of meanings 
``concept: agent: autonomous being: organization: 
educational organization:  an organization to provide education, called a school: a school at which a person was or is a student'' as the category number.

Some results of a meaning sort using the EDR dictionary are 
shown in Table \ref{tab:EDR_last_rei}\footnote{This table 
was obtained by using a Japanese dictionary. 
In the table, ``a temple'' and ``a prefecture'' belong to 
the category ``human being.'' In Japanese, 
``a temple'' and ``a prefecture'' have many meanings, 
including ``human being.''}. 
We used the first three definition terms 
as division markers.

The above analysis demonstrates that 
a meaning sort can be done using any {\sf is-a} thesaurus. 
However, 
there is a problem in that 
the order of the branching-point nodes of a hierarchical structure 
is ambiguous. 
In the case shown in Table \ref{tab:EDR_last_rei}, 
the order is 
the alphabetical order of the strings in the definition terms. 
It is better to specify the order manually, 
but if this is too difficult, it is better to 
do a meaning sort of the definition terms themselves 
by using another dictionary or thesaurus, e.g. the BGH thesaurus. 

\subsection{Msort 
using a dictionary where 
each word is expressed with a set of multiple features}
\label{sec:hukusuu_zokusei}

In some dictionaries, 
each word is expressed with a set of multiple features \cite{mental_model} \cite{Taylor89}. 
For example, 
the research of the IPAL Japanese generative dictionary \cite{ipalg98e} 
gives multiple features 
to various words having the meaning of the containers 
in Table \ref{tab:ipal_hukusuu_zokusei_rei}. 
In this table, 
''---'' means that 
the feature value is not specified.

\begin{table}[t]
  \caption{Example of a dictionary in which 
each word is assigned multiple features}
  \label{tab:ipal_hukusuu_zokusei_rei}
  \begin{center}
\begin{tabular}[c]{|l|ccccc|}\hline
Word   & \multicolumn{5}{c|}{Feature}\\\cline{2-6}
        & \multicolumn{1}{c}{\shurui}& \multicolumn{1}{c}{\taishoubutsu} & \multicolumn{1}{c}{\keijou} & \multicolumn{1}{c}{\saizu} & \multicolumn{1}{c|}{\zaishitsu}\\\hline
\hirautsuwa  & --- & --- & --- & --- & ---\\
\ishiwan      & \wakeishiki  & --- & \hukai  & --- & \toujikeishiki\\
\kiwan      & \wakeishiki  & --- & \hukai  & --- & \mokuseino  \\
\yunomi  & \wakeishiki  & \ryokutya & \hukai  & --- & \toujikeishiki \\
\sara      & --- & --- & \asai  & --- & ---\\\hline
\end{tabular}
\end{center}
\end{table}

\begin{table}[t]
  \caption{Result of an Msort, from the leftmost feature}
  \label{tab:ipal_hukusuu_zokusei_rei_hidari}
  \begin{center}
\begin{tabular}[c]{|l|ccccc|}\hline
Word    & \multicolumn{5}{c|}{Feature}\\\cline{2-6}
        & \multicolumn{1}{c}{\shurui}& \multicolumn{1}{c}{\taishoubutsu} & \multicolumn{1}{c}{\keijou} & \multicolumn{1}{c}{\saizu} & \multicolumn{1}{c|}{\zaishitsu}\\\hline
\hirautsuwa  & --- & --- & --- & --- & ---\\
\sara      & --- & --- & \asai  & --- & ---\\
\ishiwan      & \wakeishiki  & --- & \hukai  & --- & \toujikeishiki\\
\kiwan      & \wakeishiki  & --- & \hukai  & --- & \mokuseino  \\
\yunomi  & \wakeishiki  & \ryokutya & \hukai  & --- & \toujikeishiki \\\hline
\end{tabular}
\end{center}
\end{table}

It is possible to do an Msort in the case of 
such a dictionary. 
We have only to treat the information as if 
each feature is equivalent to a level 
in an imaginary `{\sf is-a}' thesaurus. 
In Table \ref{tab:ipal_hukusuu_zokusei_rei}, 
if we assume that 
the features, from left to right, correspond to 
the levels, from top to bottom, of an imaginary thesaurus, 
the levels become {\it \shurui, \taishoubutsu, \keijou, \saizu}, \ and {\it \zaishitsu}, 
and 
a category number represents 
{\it \shurui:\taishoubutsu:\keijou:\saizu:\zaishitsu}, 
which is essentially the same situation as for the EDR data. 
For example, 
the category number of {\it wan2} ({\it a wooden bowl}) is 
{\it \wakeishiki: ---: \hukai: ---: \mokuseino}. 
(Actually in order to do an Msort of 
feature values, we may change 
{\it \wakeishiki, \hukai}, and {\it \mokuseino} into the corresponding 
category numbers in BGH.) 
We simply do an Msort, 
assuming that each word has such a category word. 
The result of this Msort is shown in 
Table \ref{tab:ipal_hukusuu_zokusei_rei_hidari}. 

\begin{table}[t]
  \caption{The result of an Msort, from the rightmost feature}
  \label{tab:ipal_hukusuu_zokusei_rei_migi}
  \begin{center}
\begin{tabular}[c]{|l|ccccc|}\hline
Word    & \multicolumn{5}{c|}{Feature}\\\cline{2-6}
        & \multicolumn{1}{c}{\shurui}& \multicolumn{1}{c}{\taishoubutsu} & \multicolumn{1}{c}{\keijou} & \multicolumn{1}{c}{\saizu} & \multicolumn{1}{c|}{\zaishitsu}\\\hline
\hirautsuwa  & --- & --- & --- & --- & ---\\
\sara      & --- & --- & \asai  & --- & ---\\
\ishiwan      & \wakeishiki  & --- & \hukai  & --- & \toujikeishiki\\
\yunomi  & \wakeishiki  & \ryokutya & \hukai  & --- & \toujikeishiki \\
\kiwan      & \wakeishiki  & --- & \hukai  & --- & \mokuseino  \\\hline
\end{tabular}
\end{center}
\end{table}

Table \ref{tab:ipal_hukusuu_zokusei_rei_hidari} 
shows the result of an Msort based on the 
supposition that the leftmost feature is the most important. 
Which feature is most important is, in fact, not clear. 
For example, 
if we suppose that the rightmost feature is the most important 
and we do an Msort from that feature, 
we get a different result, 
as shown in Table \ref{tab:ipal_hukusuu_zokusei_rei_migi}. 
From a dictionary with multiple features, 
we can get various results of Msorts in this wasy, 
by changing the features which are thought to be most important. 
This means that 
users can do an Msort 
in any order of features that they may be interested in. 
This kind of dictionary, that is, the kind 
which provides multiple features, is therefore very flexible. 

\begin{figure}[t]
  \begin{center}
    \begin{minipage}{7cm}
      \begin{center}
	\leavevmode
      \epsfile{file=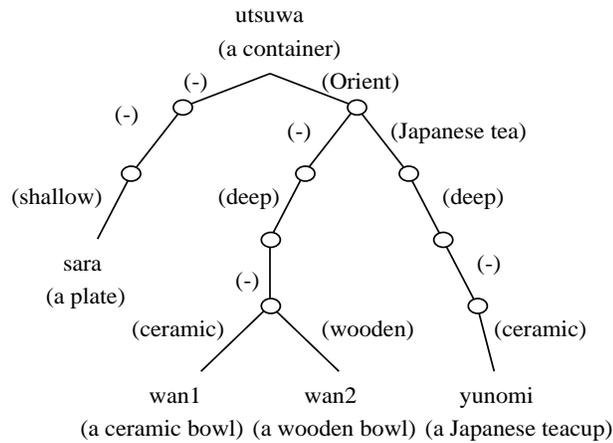,height=6cm,width=9cm} 
      \end{center}
    \caption{Hierarchical thesaurus of meaning sort from the leftmost feature}
    \label{fig:ipal_hukusuu_zokusei_rei_hidari}
    \end{minipage}
  \end{center}
\end{figure}

\begin{figure}[t]
  \begin{center}
    \begin{minipage}{7cm}
      \begin{center}
	\leavevmode
        \epsfile{file=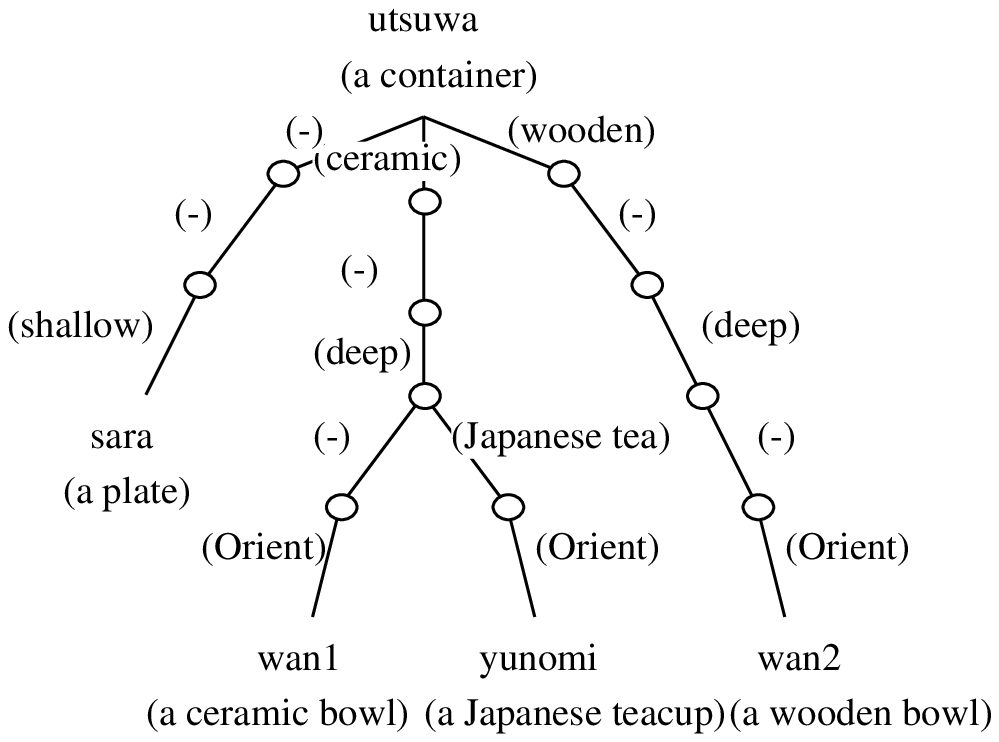,height=6cm,width=9cm} 
      \end{center}
    \caption{Hierarchical thesaurus of meaning sort from the rightmost feature}
    \label{fig:ipal_hukusuu_zokusei_rei_migi}
    \end{minipage}
  \end{center}
\end{figure}

When a hierarchical thesaurus is used to 
examine this, 
there are further interesting results. 
We can assume that 
each feature corresponds to a level of the hierarchical thesaurus, 
so we can construct many kinds of hierarchical thesauri 
by changing the correspondence between levels and features.  
For example, 
we can construct the hierarchical thesaurus 
shown in Figure \ref{fig:ipal_hukusuu_zokusei_rei_hidari} 
from the result of an Msort from the leftmost feature as shown in 
Table \ref{tab:ipal_hukusuu_zokusei_rei_hidari}. 
We can construct a hierarchical thesaurus 
shown in Figure \ref{fig:ipal_hukusuu_zokusei_rei_migi} 
from the result of an Msort from the rightmost feature as 
shown in Table \ref{tab:ipal_hukusuu_zokusei_rei_migi}. 
In the thesaurus of Figure \ref{fig:ipal_hukusuu_zokusei_rei_hidari}, 
we can see the semanitical similarity 
between {\it wan1} ({\it a ceramic bowl}) and {\it wan2} ({\it a wooden bowl}). 
In the thesaurus of Figure \ref{fig:ipal_hukusuu_zokusei_rei_migi}, 
we can understand that 
{\it wan1} ({\it a ceramic bowl}) and {\it wan2} ({\it a Japanese teacup}) 
are semantically similar 
in that they are both \toujikeishiki .
Such construction of multiple thesauri has led 
to further research into a multi-dimensional thesaurus. 
The necessity for a multi-dimensional thesaurus was discussed 
in Kawamura's paper \cite{miyazaki94Ae}. 
Kawamura's paper argued that 
if we divide {\it a bird} and {\it an airplane} 
into other categories at a relatively higher level of a hierarchy 
than the level at which entries are divided according to 
whether the item can fly or not, 
we will not be able to see that 
{\it a bird} and {\it an airplane} are semantically similar 
in that they can both fly. 
Therefore 
a dictionary with multiple features, 
which can be flexibly reconfigured into 
hierarchical thesauri of many kinds, 
would be very useful, 
and the construction of such a dictionary 
is necessary for reasons of practicality. 
Also, we have our doubts as to whether 
it is necessary to make a word dictionary 
in the form of a hierarchical thesaurus. 
Looking at Table \ref{tab:ipal_hukusuu_zokusei_rei}, 
because all the features of {\it utsuwa} ({\it a container}) are 
``---'' representing no specification of feature values, 
we are able to see that 
{\it utsuwa} ({\it a container}) is super-ordinate to the other words 
by looking at the information on the multiple features. 
We can estimate 
super-ordinate and subordinate relation 
from the inclusion relationships of features, 
so construction of a hierarchical thesaurus as such is not necessary. 
A dictionary with multiple features is all that is necessary. 
Furthermore, a dictionary with multiple features has 
a further advantage in that 
we can define the similarity of two words in terms of 
the proportion of features that are the same for both words. 
Although 
a high-order predicative logic and 
a natural language sentence can be 
thought of as the true semantic descriptions of words, 
we think that 
a dictionary using multiple features would be useful 
in that it can be handled by 
existing natural language processing techniques, and 
can handle various multi-dimensional thesauri. 

If such a dictionary is constructed, 
it would be convenient for meaning sort, 
since it would allow users to do interest-based meaning sort.

\section{Three examples of using an Msort}
\label{sec:riyourei}

In this section, 
we describe three major applications for which 
an Msort is useful. 

\subsection{Dictionary construction}

\begin{table}[p]
  \caption{Example construction of a case frame of the verb {\it eat}}
  \label{tab:taberu_case_frame}
  \begin{center}
(a) Results of an Msort of terms in the nominative case

\begin{tabular}[c]{|l@{\hspace{0.75cm}}p{9cm}|}\hline
(\doubutsu) &  cattle, a calf, fish\\
(\ningen) &  we, us, all, myself, babies, a parent, a sister, a customer, a Japanese, a nurse, a writer\\\hline
\end{tabular}

\vspace{0.3cm}

(b) Results of an Msort of terms in the objective case

\begin{tabular}[c]{|lp{9cm}|}\hline
(\doubutsu) &  an animal, shellfish, plankton\\
(\seisanbutsu) & prey, a product, a material, food, feed, Japanese food, Japanese-style food, Western food, Chinese food, a rice ball, gruel, sushi, Chinese noodles, macaroni, sandwiches, a pizza, a steak, a barbecued dish, tempura, fried food, cereals, rice, white rice, Japanese rice, barley, kimchi, sugar, jam, a confection, a cake, a cookie, ice cream \\
(Body part) &  the mortal remains, the liver\\
(\shokubutsu) &  a gene, a plant, grass, a sweet pepper, chicory, a mulberry, a banana, a matsutake mushroom, kombu\\
(\genshou) &  a delicate flavor, snow \\
(\kankei) &  the content\\
(Activity) &  breakfast, lunch, dinner, supper \\\hline
\end{tabular}

\vspace{0.3cm}

(c) Results of an Msort of terms in the optional cases 

\hspace*{-0.2cm}
\mbox{(In Japanese, ``in'', ``on'', and ``by'' are expressed by}

\mbox{the same word, so, we cannot divide data according to ``in'' or ``by'')}

\begin{tabular}[c]{|lp{9cm}|}\hline
(\ningen) &  (by) myself \\
(\soshiki) &  (in) an office, (in) a restaurant, (in) a hotel \\
(\seisanbutsu) & (by) soy sauce, (in) a dressing room, (in) bed, (on) a table\\
(\kuukan) &  (on) the spot, (in) the whole area, (on) a train \\
(\suuryou) &  (by) two persons, (at) a rate, (by) many people \\
(Activity) & (at) work, (in) a meeting\\\hline
\end{tabular}

\end{center}
\end{table}

Table \ref{tab:taberu_case_frame} shows 
the construction of a case frame for 
the verb {\it eat} according to data in a noun-verb relational dictionary 
as an example. 
The table shows the results 
of an Msort of NPs which may be 
taken as case elements of {\it eat}. 
It is easy 
to manually construct a case-frame dictionary 
from such data, as shown in Table \ref{tab:taberu_case_frame}. 
The nominative case of {\it eat} consists of agents, such as 
animals and people, and 
the objective case consists of various NPs mainly meaning foods. 
Regarding the optional case, 
various phrases such as 
{\it by myself}, {\it in an office}, and {\it in a meeting} 
are also included. 

The construction of 
a verbal case-frame dictionary 
is one example of the potential applications of an Msort. 
A similar construction process can also be easily applied to copulas 
and other kinds of relationships among words. 
An Msort is not only useful for constructing dictionaries, 
but also for examining data and 
extracting important information in language investigation. 
An Msort is also useful for examining data 
in the process of knowledge acquisition. 

\subsection{Tagged corpus construction (related to semantic similarity)}

Recently, various corpora have been under construction 
\cite{Marcus93,EDR93e,kurohashi:nlp97_e,rwce}, 
and the investigation of corpus-based learning algorithms 
is attracting much attention \cite{murata_coling2000}. 
In this section, we demonstrate how an Msort can be useful 
in the construction of corpora. 

Suppose that 
we want to disambiguate the meanings of  {\it of} 
in \Noun{X} {\it of} \Noun{Y} 
by using the example-based method \cite{Nagao_EBMT}. 
In this case, we need a tagged corpus for 
semantic analysis of the noun phrases in \Noun{X} {\it of} \Noun{Y}. 
We attach 
semantic relationships such as {\it Part-of} and {\it Location} 
to each example of the noun phrases. 
When we do an Msort of 
these phrases, 
similar examples are grouped together 
and the tagging of semantic relationships by hand is made easier. 

\begin{table}[t]
  \caption{Construction of a manually tagged corpus for the semantic analysis of 
noun phrases in ``\Noun{X} of \Noun{Y}''}
  \label{tab:make_corpus}
  \begin{center}
\begin{tabular}[c]{|l|l|l|}\hline
  \Noun{X}&  \Noun{Y}    & Semantic Relation \\\hline
  an affair    &  Panama      &      Location     \\
  an affair    &  a junior high school     &      Location      \\
  an affair    &  an army         &      Location     \\
  an affair    &  an album    &      Indirect-determiner \\
  an affair    &  a tanker    &      Indirect-determiner \\
  an affair    &  the worst   &      Adjective-feature  \\
  an affair    &  the largest &      Adjective-feature  \\
  a property   &  the circumference     &      Location     \\
  items    &  both countries  &      Object-agent  \\
  items   &  documentary records   &      Field-determiner  \\
  items   &  a general meeting      &      Object-agent  \\
  a provision    & the Upper House &      Field-determiner\\
  a provision    &  a new law   &      Field-determiner\\
  a provision    &  a treaty    &      Field-determiner\\
  a provision    &  an agreement    &      Field-determiner  \\\hline
\end{tabular}
\end{center}
\end{table}

Table \ref{tab:make_corpus} shows 
part of a manually tagged corpus. 
In this example 
we have supposed that \Noun{X} in \Noun{X} {\it of} \Noun{Y} will 
be the more important NP, 
so we first did an Msort of \Noun{X}, 
and then did one of \Noun{Y}. 
Although the technical terms representing the semantic relationships 
in the table are specialized, 
it can be seen that 
the examples which are grouped together by this Msort often have 
the same relationship. 
Also, when semantically similar examples are grouped together like this, 
the cost of tagging is decreased. 

In the example-based method, 
the tag attached to the example that 
is the most similar to the input phrase 
is judged to be the result of the analysis. 
An Msort performs the function of 
grouping similar examples. 
The example-based method and the Msort 
both use word similarity, and this 
is an advantage of both techniques. 

In this section, 
we noted that 
using the Msort is an efficient way 
to construct a noun-phrase corpus. 
In addition, when a certain corpus uses words, 
we can also use an Msort for the construct of it. 

\subsection{Information retrieval}

Information retrieval activity has increased with 
the growth of the Internet. 
An Msort can also easily be applied to this area. 

For example, in research conducted by 
Tsuda and Senda, 
the features of a document database were displayed 
to users by using multiple keywords \cite{tsuda94Ae}. 
For example, 
assume that the document database we want to display has 
the following set of keywords. 

\begin{center}
\fbox{
  \begin{minipage}[h]{8cm}
retrieval, a word, a document, construction, candidate, a number, a keyword
  \end{minipage}
}
\end{center}

Displaying the list of words in a random order is 
not very convenient for users. 
However, if we do an Msort of the keywords, 
we can obtain the following list: \\

\begin{center}
{
\begin{tabular}[h]{ll}
(\suuryou)      & a number \\
(\tyuushoukankei)  & candidate \\
(\ningenkatsudou)  & retrieval \\
            & a document, a keyword, a word,\\
            & construction \\
\end{tabular}\\
}
\end{center}

(Here, we have displayed words with 
the same first three-digit BGH category number 
on the same line.) 
This method provides a more useful perspective for users. 

In some cases 
we may display many keywords and 
ask the users to select the appropriate ones \cite{tsuda94Ae}. 
In such a case, 
if we do not have another way of arranging the words 
in an appropriate order, 
it is convenient for users if we use an Msort. 

\section{Conclusion}

In summary, we have 
introduced a useful method of arranging words semantically and 
shown how to implement it by using thesauri. 
We gave three major examples of the applications of an Msort 
(dictionary construction, tagged corpus construction, 
and information presentation). 

Since there is no doubt that 
a word list in a meaning-order is easier to use 
than a word list in a random-order, 
the Msort, which can easily produce 
a word list in a meaning-order, 
must be useful and effective.  

The Msort is a very useful tool for natural language processing, 
and NLP research can be made more efficient by applying it.

\bibliographystyle{plain}
\bibliography{mysubmit}

\begin{thebibliography}{10}

\bibitem{EDR93e}
EDR.
\newblock {\em EDR Electronic Dictionary Technical Guide}.
\newblock EDR (Japan Electronic Dictionary Research Institute, Ltd.), 1993.

\bibitem{IPALe}
IPA.
\newblock ({Information--Technology Promotion Agency, Japan}). {{\em IPA
  Lexicon of the Japanese Language for Computers IPAL (Basic Verbs)}}, 1987.
\newblock (in Japanese).

\bibitem{miyazaki94Ae}
Kazumi Kawamura and Masahiro Miyazaki.
\newblock Multi-dimensional thesaurus with various facets.
\newblock In {\em Information Processing Society of Japan, the 48th National
  Convention, 3Q-2}, pages 75--76, 1994.
\newblock (in Japanese).

\bibitem{kurohashi:nlp97_e}
Sadao Kurohashi and Makoto Nagao.
\newblock {Kyoto University text corpus project}.
\newblock pages 115--118, 1997.
\newblock (in Japanese).

\bibitem{mental_model}
P.N.~Johnson Laird.
\newblock {\em Mental models}.
\newblock Cambridge Univ. Press, 1983.

\bibitem{Marcus93}
Mitchell~P. Marcus, Beatrice Santorini, and Mary~Ann Marcinkiewicz.
\newblock Building a large annotated corpus of {English}: the {Penn Treebank}.
\newblock {\em Computational Linguistics}, 19(2):310--330, 1993.

\bibitem{ipalg98e}
Kenichi Murata, Naoko Ishida, Ryoya Okabe, Masaki Hosoi, Wakako Kashino, and
  Hajime Iizuka.
\newblock {R\&D for IPAL (SURFACE/DEEP)}.
\newblock {\em Proceedings of the 17th IPA Technological Meeting}, pages
  149--158, 1998.
\newblock (in Japanese).

\bibitem{murata_coling2000}
Masaki Murata, Kiyotaka Uchimoto, Qing Ma, and Hitoshi Isahara.
\newblock Bunsetsu identification using category-exclusive rules.
\newblock In {\em COLING 2000}, pages 565--571, 2000.

\bibitem{Nagao_EBMT}
Makoto Nagao.
\newblock {A Framework of a Mechanical Translation between Japanese and English
  by Analogy Principle}.
\newblock {\em {\em Artificial and Human Intelligence\/}}, pages 173--180,
  1984.

\bibitem{NLRI64ae}
NLRI.
\newblock {\em {\rm (National Language Research Institute)}. Word List by
  Semantic Principles}.
\newblock Syuei Syuppan, 1964.
\newblock (in Japanese).

\bibitem{rwce}
RWC.
\newblock {RWC} text database, second edition.
\newblock (in Japanese), 1998.

\bibitem{Taylor89}
John~R. Taylor.
\newblock {\em Linguistic Categorization}.
\newblock Oxford University Press, 1989.

\bibitem{tsuda94Ae}
Koji Tsuda and Shuuji Senda.
\newblock Query formulation support by automatically generated term-to-term
  links.
\newblock In {\em Information Processing Society of Japan, the 48th National
  Convention, 4E-6}, pages 157--158, 1994.
\newblock (in Japanese).

\end{thebibliography}

\end{document}